\documentclass[10pt,twocolumn,letterpaper]{article}

\usepackage{iccv}
\usepackage{times}
\usepackage{epsfig}
\usepackage{graphicx}
\usepackage{amsmath}
\usepackage{amssymb}

\usepackage{multirow}
\usepackage{adjustbox}
\usepackage{breakurl}

\usepackage[breaklinks=true,bookmarks=false]{hyperref}

\iccvfinalcopy % *** Uncomment this line for the final submission

\def\iccvPaperID{1039} % *** Enter the ICCV Paper ID here

\ificcvfinal\fi

\begin{document}

%%%%%%%%% TITLE
\title{Controllable Video Captioning  with POS Sequence Guidance \\Based on Gated Fusion Network}

\author{Bairui Wang$^1$\thanks{This work was done while Bairui Wang was a Research Intern with Tencent AI Lab.} \qquad Lin Ma$^2$\thanks{Corresponding authors.} \qquad Wei Zhang$^{1\dagger}$ \qquad Wenhao Jiang$^2$  \qquad Jingwen Wang$^2$ \qquad Wei Liu$^2$  \\
$^1$School of Control Science and Engineering, Shandong University \qquad $^2$Tencent AI Lab \\
{\tt\small \{bairuiwong, forest.linma, cswhjiang, jaywongjaywong\}@gmail.com} \\ {\tt\small  davidzhang@sdu.edu.cn \qquad wl2223@columbia.edu}
}
% Bairui Wang (Shandong University) <bairuiwong@gmail.com>
% Lin Ma (Tencent AI Lab) <forest.linma@gmail.com>
% Wei Zhang (Shandong University) <davidzhang@sdu.edu.cn>
% Wenhao Jiang (Tencent AI Lab) <cswhjiang@gmail.com>
% Jingwen Wang (Tencent AI Lab) <jwongwang@tencent.com>
% Wei Liu (Tencent) <wl2223@columbia.edu>

% \and

\maketitle
\ificcvfinal\thispagestyle{empty}\fi

%%%%%%%%% ABSTRACT
\begin{abstract}
%   In this paper, we propose to guide the video caption generation with Part-of-Speech (POS) information, based on the gated fusion of the multiple representations of input videos. We propose a novel gated fusion network, with one particularly designed cross gating (CG) block, to effectively encode and fuse the different types of representations, \textit{e.g.}, the motion and content features of the input video. One POS sequence generator relies on the fused representation to predict the global syntactic structure, which is thereafter used to guide the video captioning generation and control the syntax of generated sentence. Specifically, a gating strategy is proposed to dynamically and adaptively incorporate the global syntactic POS information into the decoder for generating each word. Experimental results on two benchmark datasets, namely MSR-VTT and MSVD, demonstrate that our proposed model can well exploit complementary information from multiple representations, resulting in improved performances. Moreover, the generated global POS information can well capture the global syntactic structure of the sentence, and thereby can be used to control the syntactic structure of the description, which thereby not only boosts the video captioning performances but also improves the diversity of the generated captions.
    In this paper, we propose to guide the video caption generation with Part-of-Speech (POS) information, based on a gated fusion of multiple representations of input videos. We construct a novel gated fusion network, with one particularly designed cross-gating (CG) block, to effectively encode and fuse different types of representations, e.g., the motion and content features of an input video. One POS sequence generator relies on this fused representation to predict the global syntactic structure, which is thereafter leveraged to guide the video captioning generation and control the syntax of the generated sentence. Specifically, a gating strategy is proposed to dynamically and adaptively incorporate the global syntactic POS information into the decoder for generating each word. Experimental results on two benchmark datasets, namely MSR-VTT and MSVD, demonstrate that the proposed model can well exploit complementary information from multiple representations, resulting in improved performances. Moreover, the generated global POS information can well capture the global syntactic structure of the sentence, and thus be exploited to control the syntactic structure of the description. Such POS information not only boosts the video captioning performance but also improves the diversity of the generated captions. Our code is at: {\color{blue} \url{{https://github.com/vsislab/Controllable_XGating}}}.
\end{abstract}

%%%%%%%%% BODY TEXT
\vspace{-10pt}
\section{Introduction}
\begin{figure}
\centering
\includegraphics[scale=0.75]{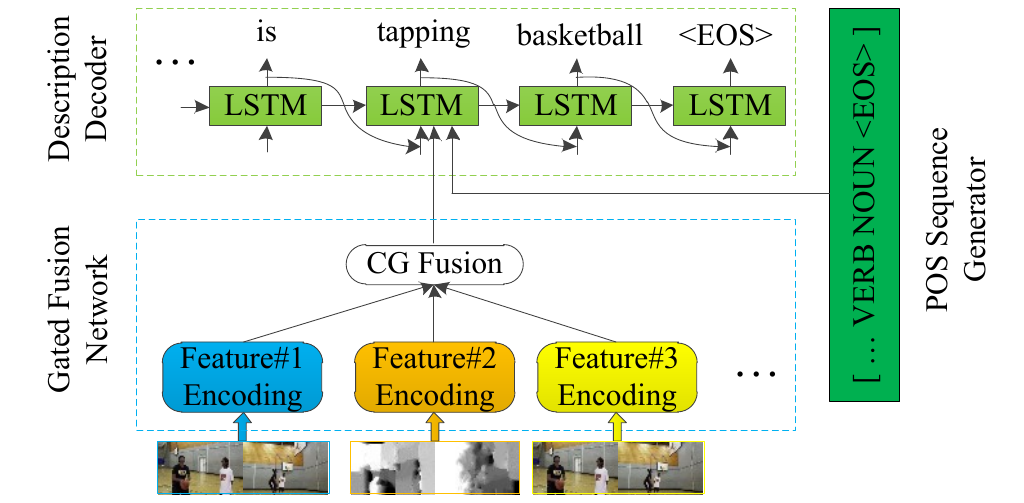} %width=\hsize
\caption{The proposed model for video captioning consists of a gated fusion network, a POS sequence generator, and a description generator. The gated fusion network extracts diverse features from videos, encodes, and fuses them together to generate a more representative video feature. Relying on the global syntactic POS information generated from the POS sequence generator and the fused video feature, the description generator produces one sentence describing the video content.}
\label{fig:the_first_fig}
\vspace{-15pt}
\end{figure}

Video captioning~\cite{jin2016describing,Zhang2019reconstruct,wang2018bidirectional} aims at automatically describing rich content in videos with natural language, which is a meaningful but challenging task for bridging vision and language.
This task can be applied for high-level video understanding in a variety of practical applications, such as visual retrieval~\cite{ma2015multimodal,song2018quantization,wang2018survey,ma2019matching},  visual question answering~\cite{ma2016learning,das2018embodied}, and so on.%auxiliary aid for visually impaired people~\cite{voykinska2016blind}.
%Recently, video captioning has received an increasing interest in both the computer vision and natural language processing communities.
Video captioning is related to image captioning which describes an image with a sentence, as a video can be regarded as a sequence of images.
However, what makes video captioning more challenging than image captioning~\cite{feng2019unsupervised,wang2019hierarchical,chen2018regularizing,jiang2018recurrent} is not only that the input of video captioning are multiple images, but also that video contains richer semantics, such as spatial/temporal information, content/motion information, and even speech information.
Obviously, the existing approahces with one single kind  feature~\cite{donahue2015long,venugopalan2014translating,wang2018reconstruction,wang2018bidirectional} are hard to comprehensively exploit the semantic meaning of a video.

% The existing approaches rely on an encoder-decoder architecture for video captioning~\cite{donahue2015long,venugopalan2014translating,venugopalan2015sequence,yao2015describing,pan2016jointly,pan2017video,gao2017video,song2017hierarchical,liu2017video,wang2018reconstruction,wang2018bidirectional}. {venugopalan2015sequence,yao2015describing,pan2016jointly,pan2017video}
% Specifically, one video is fed into an encoder to yield its representation. And the decoder, working on the encoded representation, is responsible to generate a sentence, expressing the semantic meaning of the video.

Recently, researches on describing videos from diverse representations, such as Inception\_ResNet\_V2~\cite{szegedy2017inception}, C3D~\cite{tran2015learning}, and I3D~\cite{carreira2017quo}, have proved that multiple features can improve the video captioning models~\cite{venugopalan2015sequence,yao2015describing,pan2016jointly,pan2017video}. It is reasonable as different features can capture video semantic information from different perspectives.
However, to the best of our knowledge, the existing methods simply concatenate different representations together, while neglect the relationships among them, which play an important role in fully  characterizing the video semantic meaning.

Prior video captioning methods also neglect the syntactic structure of a sentence during the generation process. Analogic to the fact that words are the basic composition of a sentence, the  part-of-speech (POS)~\cite{deshpande2018diverse} information of each word in a sentence is the basic structure of the grammar. Therefore, the POS information of the generated sentence is able to act as one prior knowledge to guide and regularize the sentence generation, if it can be obtained beforehand.
Specifically, with the obtained POS information, the decoder is aware of the POS information of the word to be generated. As such, it may help reduce the search space of the target word, which is believed to benefit the video captioning.
Besides, the changing of POS information, which can be seen as the prior knowledge of the description, is excepted to help generate sentence with more diverse syntax.

In order to fully exploit the relationships among different representations and the POS information, we propose a novel model to describe videos with POS guidance based on the gated fusion results of multiple representations, as shown in Fig.~\ref{fig:the_first_fig}. 
%\textcolor{green}{The content and motion representations are taken as the example in the figure.} 
% First, a novel gated fusion network relies on a particularly designed cross gating (CG) block to represent and fuse diverse features. Specifically, a cross gating mechanism is proposed to gate these features with respect to each other and combine them together. 
First, a novel gated fusion network relies on a particularly designed cross gating (CG) block to mutually gate diverse features with respect to each other.
As such, we can make a comprehensive representation of the video. One POS sequence generator relies on the fused video representation to yield the global POS information. Afterwards, the decoder relies on a gating strategy to dynamically and adaptively incorporate the generated global syntactic POS information for generating each word.

To summarize, the contributions of this work lie in threefold: 1) We propose a novel video captioning model, which relies on a gated fusion network incorporating multiple features information together and a POS sequence generator predicting the global syntactic POS information of the generated sentence. 2) A cross gating (CG) strategy is proposed to effectively encode and fuse different representations. The global syntactic POS information is adaptively and dynamically incorporated into the decoder to guide the decoder to produce more accurate description in terms of both syntax and semantics. 3) Extensive results on benchmark datasets indicate that the proposed fusion strategy can capture the relationships among multiple representations and descriptions with diverse syntax can be obtained by controlling the global POS sequence.
% \begin{itemize}
%     \item We propose a novel video captioning model, which relies on a gated fusion network incorporating multiple features information together and a POS sequence generator predicting the global syntactic POS information of the generated sentence.
%     \item A cross gating (CG) strategy is proposed to effectively encode and fuse different representations. The global syntactic POS information is adaptively and dynamically incorporated into the decoder to guide the decoder to produce more accurate description in terms of both syntax and semantics.
%     \item Extensive results on benchmark datasets indicate that the proposed fusion strategy can capture the relationships among multiple representations and descriptions with diverse syntax can be obtained by controlling the global POS sequence.
% \end{itemize}

\section{Related Work}

\begin{figure*}[htbp]
\centering
\includegraphics[width=0.85\hsize]{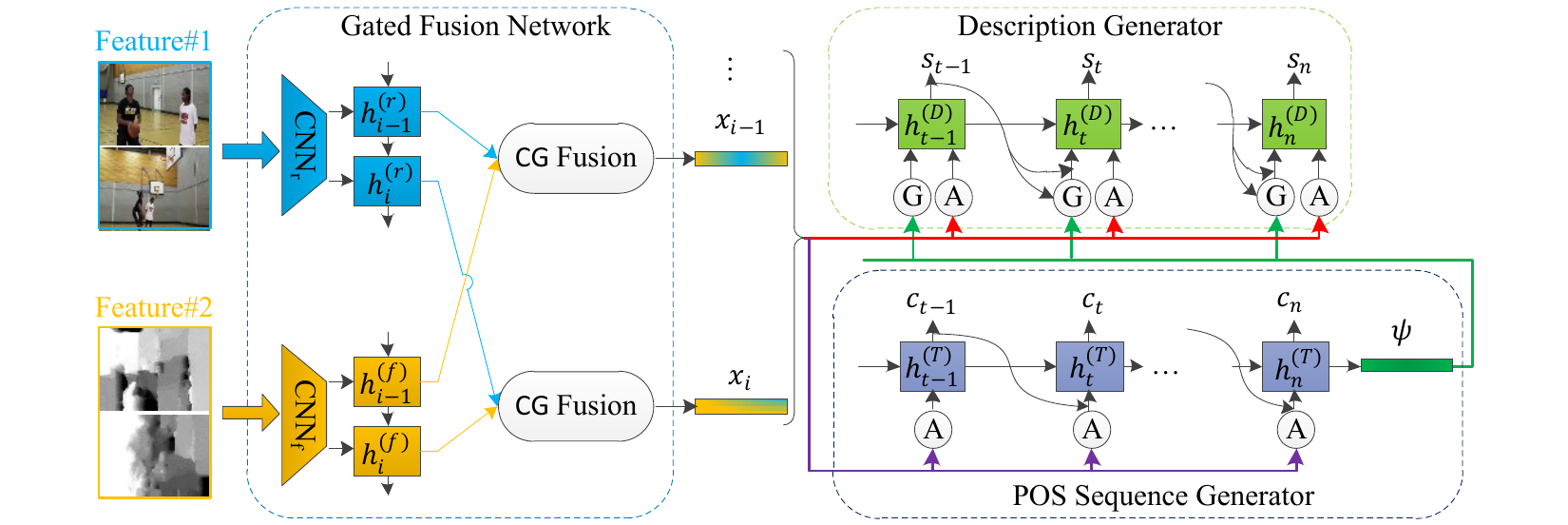}
\caption{The proposed model for video captioning consists of three components. The gated fusion network encodes and fuses multiple video representations extracted by different CNN networks. The POS generator relies on the fused video representation to predict the global syntactic POS information of the sentence to be generated. The decoder adaptively and dynamically incorporates the global POS information for generating each targeting word. \textcircled{\small{G}} denotes the cross gating mechanism and \textcircled{\small{A}} denotes the soft attention mechanism.}
\label{fig:architecture}
\vspace{-15pt}
\end{figure*}

\subsection{Video Captioning}
Previous works on video captioning adopt temporal-based methods~\cite{kojima2002natural,guadarrama2013youtube2text,rohrbach2013translating,rohrbach2014coherent,xu2015jointly}, which define a sentence template with grammar rules. The sentence is parsed into subject, verb and object, each of which is aligned with video content.
Obviously, under the predefined template with fixed syntactic structure, those methods are hard to generate flexible language descriptions.

Nowadays, benefit from the success of CNN and RNN, the sequence learning methods~\cite{venugopalan2015sequence,yao2015describing,pan2016jointly,pan2017video,chen2018less,wang2018reconstruction,wu2018interpretable} are widely used to describe video content with flexible syntactic structure.
In \cite{venugopalan2014translating}, Venugopalan~\etal obtained video representation by averaging CNN feature of each frame, which ignored the temporal information.
Compare to the average pooling, Yao~\etal and Yu~\etal employed the soft attention mechanism to dynamically summarize all frame representations~\cite{yao2015describing,yu2016video}.
Recently, to exploit more semantic information, Pan~\etal modeled the semantic-level correlation of sentence and video with a visual-semantic embedding model~\cite{pan2016jointly}.
% Wang~\etal designed a reconstructor to capture dual information between videos and captions~\cite{wang2018reconstruction}.
% Baraldi~\etal extracted the hierarchical representations from video content by detecting the temporal discontinuities.
%More recently, 
To avoid the negative impact of redundant visual information, Chen~\etal proposed a PickNet to choose key frames~\cite{chen2018less}. % instead of equal interval sampling on all frames~\cite{chen2018less}.

% More recently, diverse features that provide information in various aspects are getting more and more attention for video captioning. %among which the motion information is widely used.
%For example, 
{More recently, different features can help characterizing the video semantic meaning from different perspectives.}
Many existing works utilize the motion information~\cite{venugopalan2015sequence}, temporal information~\cite{chen2017video,jin2016describing,ramanishka2016multimodal}, and even the audio information~\cite{xu2017learning} to yield competitive performance. However, the diverse features in these works are simply concatenated with each other, which ignores the relationship among them. It is possible to further improve performance with a better fusion strategy.
% Inspired by \cite{feng2018video} where a cross gating mechanism is successfully applied to match reference videos and query videos for video re-localization,
In this paper, we design a gated fusion network to dynamically learn and highlight the correlation between different features, which is expected to fully depict and characterize the video semantic meaning.

\subsection{Captioning with POS Information}
To the best of our knowledge, the POS tag information of language description has not been introduced in the video captioning task. While in image captioning, Deshpande~\etal treated the entire POS tag sequence given by benchmark dataset as a sample, and divided them in 1024 categories by a k-medoids cluster~\cite{deshpande2018diverse}, which limits the diversity of POS sequence information.
He~\etal controlled the input of image representations based on the predefined POS tag information of each ground-truth word~\cite{he2017image}, which can hardly obtained in practical scenario.
In contrast, we predict POS sequence tag by tag, and embed them as a global POS feature to provide approximate global view on syntactic structure of the sentences.
More importantly, the syntactic structure of description is controllable by changing the POS sequence manually.

\section{Architecture}
Given a video sequence, video captioning aims to generate a natural sentence $S=\{s_1, s_2, \dots, s_n\}$ to express its semantic meaning, where $n$ denotes the length of a sentence. In this paper, we would like to make a full exploitation of the video sequence by considering diverse video features. Moreover, we also want to predict the syntactic information of the generated sentence, specifically the POS information $C=\{c_1, c_2, \dots, c_n\}$, which is thereafter leveraged for guiding the sentence generation.

We propose one model for video captioning, realized in an encoder-decoder architecture, which consists of a gated fusion network, a POS sequence generator, and a description generator, as shown in Fig.~\ref{fig:architecture}.
%The video encoder is of a two-stream architecture, with one encoding content information and the other encoding motion information.
The gated fusion network learns to exploit the relationship among different video features to make a comprehensive understanding of the video sequence.
The POS sequence generator learns to exploit the relationship between fused representation and POS tags of ground-truth descriptions, and thereby predicts global POS representation for the sentence to be generated. The description generator attentively summarizes the fused representations and generates each word by adaptively integrating the predicted global POS representation.

\subsection{Gated Fusion Network}
\label{sec:encoder_video_encoder}
Given the input videos, the gated fusion network first extracts different semantic representations for each frame by multiple CNN networks. For the convenience of expression, we take the visual content feature from RGB frames and motion feature from optical flows as examples in this section, which are denoted as $R = \{r_1, r_2, \dots, r_m \}$ and $F = \{f_1, f_2, \dots, f_m \}$, respectively, where $r_i$ and $f_i$ denote the features for the $i_{th}$ frame and optical flow of the input video, respectively.
% The $r_i$ and $f_i$ denote the different representations, specifically the content feature and the motion feature, for the $i_{th}$ frame and optical flow in the input video, respectively.
$m$ indicates the total length of the video. Based on the obtained representations $R$ and $F$, the gated fusion network performs in two stages. First,  temporal encoding of each representation is performed, respectively. Afterwards, a cross gating strategy is proposed to fuse the temporally aggregated feature together.

% Given the raw frame sequence and the corresponding dense optical flow sequence, the encoder first employs CNNs to capture the semantic information by extracting the content representation $R = \{r_1, r_2, \dots, r_m \}$ for each static frame and motion representation $F = \{f_1, f_2, \dots, f_m \}$ for the optical flow, where $r_i$, $f_i$ denote the content and motion representation of the $i_{th}$ frame and optical flow, respectively, and $m$ denotes the total length of the video. Based on the obtained representations  $R$ and $F$, the video encoder performs in two stages, with one temporally encoding the motion and content representations respectively,  and the other fusing them together.
\begin{figure}[htpb]
\vspace{-10pt}
\centering
\includegraphics[scale=0.75]{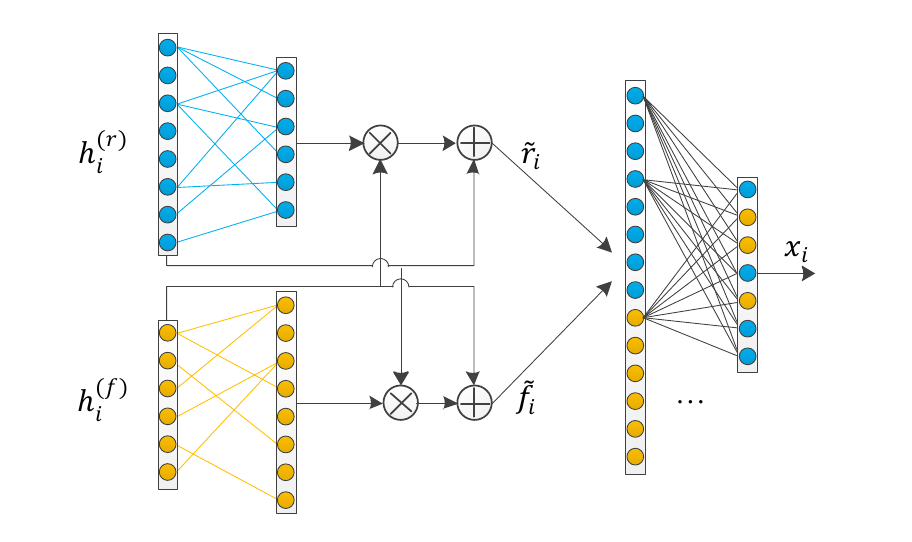}
\vspace{-3pt}
\caption{An illustration of cross gating strategy in the proposed gated fusion network. %The cross gating strategy is proposed to strengthen the information that is related to the other in multiple modality features. The residual fusion mechanism merges the high-level multi-modal features in an efficient way. 
The cross gating strategy strengthens the information that is related to each other within diverse features, and then fuses them together.
$\bigotimes$ and $\bigoplus$ denote the element-wise multiplication and addition, respectively.}
\label{fig:gated_fusion}
\vspace{-11pt}
\end{figure}

\noindent \textbf{Temporal Encoder.}
Long short term memory networks (LSTMs) are used to aggregate these representations:
% Given the high-level local representation sequence $\tilde{R}^l = \{\tilde{r}^l_1, \tilde{r}^l_2, \dots, \tilde{r}^l_m \}$ and $\tilde{F}^l = \{\tilde{f}^l_1, \tilde{f}^l_2, \dots, \tilde{f}^l_m \}$, we further employ a gated temporal encoder consisting of bi-gate function and a LSTM on them.
\begin{small}
{\setlength\abovedisplayskip{2pt}
\setlength\belowdisplayskip{2pt}
\begin{align}
\label{eq:encoder_temporal_lstm}
  \begin{split}
    % h^{(r)}_i, z^{(r,E)}_i = \text{LSTM}_r \left( \tilde{r}^l_i, h^{(r, E)}_{i-1} \right), \\
    h^{(r)}_i, z^{(r)}_i = \text{LSTM}^{(E)}_r \left( r_i, h^{(r)}_{i-1} \right), \\
    h^{(f)}_i, z^{(f)}_i = \text{LSTM}^{(E)}_f \left( f_i, h^{(f)}_{i-1} \right),
  \end{split}
\end{align}}
\end{small}
where $\text{LSTM}^{(E)}_r$ and $\text{LSTM}^{(E)}_f$ denote the LSTM units for the content and motion features, respectively. $h^{(r)}_i$, $h^{(f)}_i$, $z^{(r)}_i$ and $z^{(f)}_i$ are the corresponding hidden states and memory cells. With LSTM encoding, high-level content and motion feature sequences $\hat{R} = \{h^{(r)}_1, h^{(r)}_2, \dots, h^{(r)}_m \}$ and $\hat{F} = \{h^{(f)}_1, h^{(f)}_2, \dots, h^{(f)}_m \}$ are obtained.

\noindent \textbf{Cross Gating.}
A simple concatenation of $\hat{R}$ and $\hat{F}$ can fuse all the different features of a frame. However, such a fusion strategy ignores the relationship between these features. To take full advantage of the related semantic information, we propose a novel cross gating strategy on different features as illustrated in Fig.~\ref{fig:gated_fusion}:
\begin{small}
{\setlength\abovedisplayskip{2pt}
\setlength\belowdisplayskip{2pt}
\begin{align}
\label{eq:encoder_fusion_gate}
  \begin{split}
    \tilde{r}_i &= \text{Gating}^{(E)}_r \left ( h^{(f)}_i, h^{(r)}_i \right ), \\
    \tilde{f}_i &= \text{Gating}^{(E)}_f \left ( h^{(r)}_i, h^{(f)}_i \right ),
  \end{split}
\end{align}}
\end{small}
where $\tilde{r}_i$ and $\tilde{f}_i$ are the gated results for content and motion representations. We realize the $\text{Gating}$ function as follows:
\begin{small}
{\setlength\abovedisplayskip{2pt}
\setlength\belowdisplayskip{2pt}
\begin{align}
\label{eq:encoder_implement_gate}
  \begin{split}
    \text{Gating} \left ( x, y \right ) =  \sigma \left( wx+b \right)y + y,
  \end{split}
\end{align}}
\end{small}
where $y$ denotes the target feature, which is updated under the guidance of the driver feature $x$. $w$ and $b$ are learnable parameters, and $\sigma$ is a nonlinear activation function, which is a ReLU function in our implementation.
Obviously, in $\tilde{r}_i$ the content information related to the motion information has been strengthened by the proposed cross gating strategy. And the similar process performs on $\tilde{f}_i$, where the motion information related to content information is strengthened.

Finally, the gated representations of content and motion are fused together by a fully connected layer:
\begin{small}
{\setlength\abovedisplayskip{2pt}
\setlength\belowdisplayskip{2pt}
\begin{align}
\label{eq:encoder_late_fusion}
  \begin{split}
     x_i = w^{(E)}\left( \left[ \tilde{r}_i, \tilde{f}_i \right]+b^{(E)} \right),
  \end{split}
\end{align}}
\end{small}
where $\left[ \cdot \right]$  denotes the concatenation of inputs. $x_i$ denotes the fused representation for each frame where both content and motion information are included. $w^{(E)}$ and $b^{(E)}$ are the learnable parameters.

\subsection{POS Sequence Generator}
\label{sec:pos_generatior}
In addition to natural language descriptions, the POS of each word in sentences is also closely related to the video content. To utilize the POS information, we design a simple POS sequence generation network based on the fused representations.
Based on the fused feature sequence ${X}= \{ {x}_1, {x}_2, \dots, {x}_m \}$, the POS generator predicts POS sequence:
% \begin{align}
% \label{eq:pos_lstm}
%   \begin{split}
%     h^{(T)}_t, z^{(T)}_t &= \text{LSTM}^{(T)} \left(  \left[E(c_{t-1}),\phi_t \left( {X, h^{(T)}_{t-1}}  \right) \right],  h^{(T)}_{t-1} \right), \\
%     \psi &= h^{(T)}_n, \\
%     P(c_{t}|c_{<t},V;\theta_{pos}) &=  \text{softmax} \left( \mathbf{W}^{(T)} h^{(T)}_t + \mathbf{b}^{(T)} \right),
%   \end{split}
% \end{align}
\begin{small}
{\setlength\abovedisplayskip{2pt}
\setlength\belowdisplayskip{2pt}
\begin{equation}
\label{eq:pos_lstm}
  \begin{aligned}
    %h^{(T)}_t, z^{(T)}_t &= \text{LSTM}^{(T)} \left(  \left[E_{pos}(c_{t-1}), \right. \right. \\
    %& \left. \left.  \phi_t \left( {X, h^{(T)}_{t-1}}  \right) \right],  h^{(T)}_{t-1} \right), \\
    %\psi &= h^{(T)}_n, \\
    % h^{(T)}_t, z^{(T)}_t = \text{LSTM}^{(T)} \left(  \left[E_{pos}(c_{t-1}), \phi_t \left( {X, h^{(T)}_{t-1}}  \right) \right],  h^{(T)}_{t-1} \right), \\
    % \text{temp} &= \left[E_{pos}(c_{t-1}), \phi_t \left( {X, h^{(T)}_{t-1}}  \right) \right], \\
    h^{(T)}_t, z^{(T)}_t = \text{LSTM}^{(T)} \left( \left[E_{pos}(c_{t-1}), \phi_t \left( {X, h^{(T)}_{t-1}}  \right) \right],  h^{(T)}_{t-1} \right), \\
    P(c_{t}|c_{<t},V;\theta_{pos}) =  \text{softmax} \left( \mathbf{W}^{(T)} h^{(T)}_t + \mathbf{b}^{(T)} \right),
  \end{aligned}
\end{equation}}
\end{small}
where $h^{(T)}_t$ and $z^{(T)}_t$ are hidden state and memory cell of POS generator.
%Suppose that after $n$ time steps all the POS tags in a sentence have been predicted, then the last hidden state $h^{(T)}_n$ of the LSTM can be regarded as the global representation of the POS sequence. 
$c_{t-1}$ denotes the POS tag predicted at the previous step, $E_{pos}$ is an embedding matrix for POS tags and we denote by $E_{pos}(c_{t-1})$ the embedding vector of POS tag $c_{t-1}$. $\theta_{pos}$, $\mathbf{W}^{(T)}$ and $\mathbf{b}^{(T)}$ denote the learnable parameters in POS sequence encoder. $P(c_t|c_{<t},V;\theta_{pos})$ means the probability of predicting correct POS tag $c_t$ given the previous tags $c_{<t} = \{c_1, c_2, \dots, c_{t-1} \}$ and input video $V$.

Please note that the symbol $\phi_t \left( \cdot \right)$ in Eq.~(\ref{eq:pos_lstm}) denotes the soft attention process at time step $t$, which yields a vector representation $\phi_t\left({X, h^{(T)}_{t-1}} \right)$ with different weights on ${X}$: %$X= \{ x_1, x_2, \dots, x_m \}$:
\begin{small}
{\setlength\abovedisplayskip{2pt}
\setlength\belowdisplayskip{2pt}
\begin{align}
\label{eq:pos_atten_sum}
  \begin{split}
    \phi_t \left( X, h^{(T)}_{t-1}\right) = \sum_{i=1}^{m}{\alpha_{t,i} x_i} ,  \\
  \end{split}
\end{align}}
\end{small}
where $\sum_{i=1}^{m}{\alpha_{t,i}}=1$ and $\alpha_{t,i}$ denotes the attention weights computed for the $i_{th}$ fused representation at the $t_{th}$ time step. It encourages the POS sequence encoder to select the useful information related with the POS tag predicted at the current step. The attentive weight $\alpha$ is computed by:
\begin{small}
{\setlength\abovedisplayskip{2pt}
\setlength\belowdisplayskip{2pt}
\begin{align}
\label{eq:pos_atten_weights}
  \begin{split}
    e_{t,i} &= \mathbf{w}^{(T)\top} \text{tanh} \left( \mathbf{W}^{(T)} h^{(T)}_{t-1} + \mathbf{U}^{(T)} x_{i} + \mathbf{b}^{(T)} \right), \\
    \alpha_{t,i} &= \exp \left( e_{t,i} \right ) / \sum_{k=1}^{m}{\exp \left(e_{t,i} \right)},  
  \end{split}
\end{align}}
\end{small}
where $\mathbf{w}^{(T)\top}$, $\mathbf{W}^{(T)}$, $\mathbf{U}^{(T)}$ and $\mathbf{b}^{(T)}$ are learnable parameters.

When the prediction for the whole POS sequence finishes, the last hidden state $\psi = h^{(T)}_n$ of the LSTM is expected to capture the global information of the POS sequence of the generated sentence, which is further used to guide the description generation and control the syntactic structure to generate the sentence. %For clarity, we denotes $h_m$ as $\psi$.

\begin{figure}[tbp]
\centering
\includegraphics[width=\hsize]{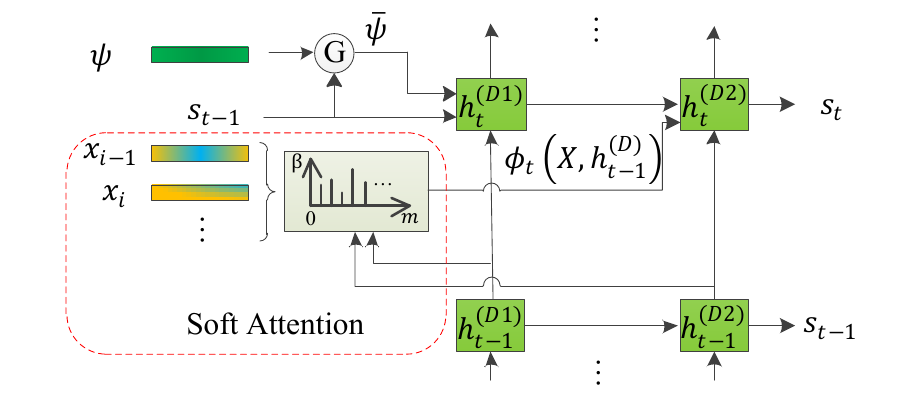}
% \vspace{-3pt}
\caption{The architecture of the proposed description generator. At each time step, the cross gating~\textcircled{\small{G}} is performed on the embedding vector of predicted word and the predicted global POS feature, by which the POS information related to the current word is dynamically and adaptively incorporated. The soft attention mechanism dynamically summarizes the fused frame features.}
\label{fig:decoder}
\vspace{-10pt}
\end{figure}

\subsection{Description Generator}
The description generator produces sentence description for video based on the fused video representation $X=\{x_1, x_2, \dots, x_m\}$ learned by the video encoder and the predicted global POS representation $\psi$.
We employ a hierarchical decoder, which consists of a two-layer LSTM. The first layer is fed with the generated word $s_{t-1}$ and the global POS feature $\psi$, while the second one takes the hidden state of first layer and an attentive summary of $X$ as input.

When describing a video $V$, the word embedding vector is updated by performing the cross gating on the global POS feature $\psi$ generated in Sec.~\ref{sec:pos_generatior}:
\begin{small}
{\setlength\abovedisplayskip{2pt}
\setlength\belowdisplayskip{2pt}
\begin{align}
\label{eq:decoder_pos_gated}
  \begin{split}
    \bar{\psi} = \text{Gating}^{(D)} \left ( E_{word}(s_{t-1}), \psi \right ), 
  \end{split}
\end{align}}
\end{small}
where $E_{word}(s_{t-1})$ is the word embedding vector of the word $s_{t-1}$, which is generated at the previous time step. As such, the global POS information with respect to the predicted word is strengthened.

Afterwards, the process of description generator is as follows:
% \begin{small}
% \begin{align}
% \label{eq:decoder_lstm}
%     \centering
%   \begin{split}
%     \phi_t(X, h^{(D)}_{t-1}) = \sum_{i=1}^{m}{\beta_{t,i} x_i }, \\
%     h^{(D1)}_{t}, z^{(D1)}_{t} = \text{LSTM}^{(D1)} \left( \left[ E_{word}(s_{t-1}), \bar{\psi} \right], h^{(D1)}_{t-1} \right), \\
%     h^{(D2)}_{t}, z^{(D2)}_{t} = \text{LSTM}^{(D2)} \left( \left[ h^{(D1)}_{t}, \phi_t(X, h^{(D)}_{t-1}) \right], h^{(D2)}_{t-1}  \right), \\
%     P( s_t | s_{<t}, V; \theta_{gen}) = \text{softmax} \left( \mathbf{W}^{(D)}_s h^{(D2)}_{t} + \mathbf{b}^{(D)}_s \right),
%   \end{split}
% \end{align}
% \end{small}
\begin{small}
{\setlength\abovedisplayskip{2pt}
\setlength\belowdisplayskip{2pt}
\begin{align}
\label{eq:decoder_lstm}
    \centering
  \begin{split}
    \phi_t(X, h^{(D)}_{t-1}) &= \sum_{i=1}^{m}{\beta_{t,i} x_i }, \\
    % \text{temp1} &= \left[ E_{word}(s_{t-1}), \bar{\psi} \right], \\
    % \text{temp2} &= \left[ h^{(D1)}_{t}, \phi_t(X, h^{(D)}_{t-1}) \right], \\
    h^{(D1)}_{t}, z^{(D1)}_{t} &= \text{LSTM}^{(D1)} \left( \left[ E_{word}(s_{t-1}), \bar{\psi} \right], h^{(D1)}_{t-1} \right), \\
    h^{(D2)}_{t}, z^{(D2)}_{t} &= \text{LSTM}^{(D2)} \left( \left[ h^{(D1)}_{t}, \phi_t(X, h^{(D)}_{t-1}) \right], h^{(D2)}_{t-1}  \right), \\
    P( s_t | s_{<t}, V; \theta_{gen}) &= \text{softmax} \left( \mathbf{W}^{(D)}_s h^{(D2)}_{t} + \mathbf{b}^{(D)}_s \right),
  \end{split}
\end{align}}
\end{small}
where $\text{LSTM}$ with subscripts $D1$ and $D2$ denote the LSTM units at the first and second layers in decoder. $\mathbf{W}^{(D)}_s$, $\mathbf{b}^{(D)}_s$, and $\theta_{gen}$ denote the learnable parameters in the description generator.
Once again, we apply the soft attention on $X$, as in Eq.~(\ref{eq:pos_atten_weights}), to dynamically select the high-level fused features, which is denoted as $\phi_t(X, h^{(D)}_{t-1})$.
The attentive weights $\beta$ satisfy $\sum_{i=1}^{m}{\beta_{t,i}}=1$. Please note that we use a hierarchical guidance consisting of $h^{(D1)}_{t}$ from the first layer and $h^{(D2)}_{t}$ from the second layer to drive the attention mechanism, which is obtained by:
%\begin{small}
%\begin{align}
%\label{eq:decoder_attention}
%  \begin{split}
%    e^{(D)}_{t,i} &= \mathbf{w}^{(D)\top}_a \text{tanh} \left( \mathbf{W}^{(D)}_a \left[h^{(D1)}_{t-1}, h^{(D2)}_{t-1}  \right] \right. \\
%    & \left. + \mathbf{U}^{(D)}_a x_i + \mathbf{b}^{(D)}_a \right), \\  %\bar{h}^{(E)}
%    \beta_{t,i} &= \exp \left( e^{(D)}_{t,i} \right ) / \sum_{j=1}^{m}{\exp \left(e^{(D)}_{t,j} \right) }.   \\
%  \end{split}
%\end{align}
%\end{small}
\begin{small}
{\setlength\abovedisplayskip{2pt}
\setlength\belowdisplayskip{2pt}
\begin{align}
\label{eq:decoder_attention}
  \begin{split}
    % \text{temp} &= \left[h^{(D1)}_{t-1}, h^{(D2)}_{t-1}  \right], \\
    e^{(D)}_{t,i} &= \mathbf{w}^{(D)\top}_a \text{tanh} \left( \mathbf{W}^{(D)}_a \left[h^{(D1)}_{t-1}, h^{(D2)}_{t-1}  \right] + \mathbf{U}^{(D)}_a x_i + \mathbf{b}^{(D)}_a \right), \\  %\bar{h}^{(E)}
    \beta_{t,i} &= \exp \left( e^{(D)}_{t,i} \right ) / \sum_{j=1}^{m}{\exp \left(e^{(D)}_{t,j} \right) }.   \\
  \end{split}
\end{align}}
\end{small}

\subsection{Training}
The proposed model is trained in two stages. We first freeze the parameters of the description generator, and train the POS sequence generator in a supervised way with the purpose of obtaining accurate global POS information. The loss function is defined as the negative log probability of each POS sequence:
\begin{small}
{\setlength\abovedisplayskip{2pt}
\setlength\belowdisplayskip{2pt}
\begin{align}
\label{eq:pos_total_loss}
  \begin{split}
    \mathcal{L}_{pos}\left(\theta_{pos}\right) = -\sum^{N}_{k=1}{\log P \left( C^k | V^k; \theta_{pos} \right) },
  \end{split}
\end{align}}
\end{small}
where $N$ is the total number of training data, and the probability of one POS sequence is defined as:
\begin{small}
{\setlength\abovedisplayskip{2pt}
\setlength\belowdisplayskip{2pt}
\begin{align}
\label{eq:pos_sequence_loss}
  \begin{split}
    P \left( C|V;\theta_{pos} \right) = \prod_{t=1}^{n}{P \left(c_t | c_{<t}, V; \theta_{pos} \right) }.
  \end{split}
\end{align}}
\end{small}

When the POS generator converges, we predicts the global POS information based on the POS generator for each  video sequence. Then, the video encoder and description generator are jointly trained by minimizing the cross-entropy loss, which is similar to Eq.~(\ref{eq:pos_total_loss}) and ~(\ref{eq:pos_sequence_loss}):
\begin{small}
{\setlength\abovedisplayskip{2pt}
\setlength\belowdisplayskip{2pt}
\begin{align}
\label{eq:decoder_total_loss}
  \begin{split}
    \mathcal{L}_{gen} \left(\theta_{gen}\right) = -\sum^{N}_{k=1}{\log P \left( S^k | V^k; \theta_{gen} \right) }, \\
  \end{split}
\end{align}}
\end{small}
where $P\left( S | V;\theta_{gen} \right) = \prod_{t=1}^{n}{P \left(s_t | s_{<t}, V; \theta_{gen} \right) }$.

Besides, we intend to directly train the captioning models guided by evaluation metrics, specifically the CIDEr~\cite{vedantam2015cider} in this work, instead of the cross-entropy loss. As such an evaluation metric is discrete and non-differentiable, we resort to the self-critical sequence training~\cite{rennie2017self} to further boost the performance of the proposed method.  More details about the self-critical strategy can be found in the supplementary material.

\subsection{Inference}
As the global POS information generated by the proposed POS sequence generator can help improve captioning performance and control the syntactic structures of the generated descriptions, we verify our method in two ways.
First, the POS sequence generator generates global POS information without any human intervention. %, so that we can study the performance improvement brought by the global POS sequence.
secondly, we control the global POS information by changing the predicted POS tags. For example, we change one or more predicted POS tags, based on which, the corresponding global POS information is then generated.

In both two ways, the global POS information is utilized by the description generator to predict the captions. The first verification is for demonstrating the performance improvements brought by the proposed gated fusion network and global POS sequence guidance, while the other one aims to present the controllablity for the syntactic structure of the video description generation.

\section{Experiments}
In this section, we evaluate the proposed video captioning method on Microsoft Research video to text (MSR-VTT)~\cite{xu2016msr} %, ActivityNet 1.3 datasets~\cite{caba2015activitynet}\textcolor{green}{(where is the table?)}, 
and Microsoft Video Description Corpus (MSVD)~\cite{chen2011collecting} with the widely-used metrics including BLEU@\textit{N}~\cite{papineni2002bleu}, METEOR~\cite{banerjee2005meteor}, ROUGE-L~\cite{lin2004rouge}, and CIDEr~\cite{vedantam2015cider}. They are denoted as B@\textit{N}, M, R, and C respectively, where \textit{N} varies from 1 to 4. The codes for these metrics have been released on Microsoft COCO evaluation server~\cite{chen2015microsoft}.
We first briefly describe the datasets used for evaluation, followed by the implementation details. Afterwards, we discuss the experiment results on video captioning.

\subsection{Datasets}
\textbf{MSR-VTT.}
The MSR-VTT is a large-scale dataset for video captioning, which covers the most diverse visual contents so far.
It contains 10,000 video clips from 20 categories and 200,000 video-caption pairs with 29,000 unique words in total. Each video clip corresponds to 20 English sentence descriptions.
Following the existing work, we use the public splits for training and testing, where 6,513 for training, 497 for validation, and 2,990 for testing.

% \textbf{ActivityNet 1.3.}
% The ActivityNet 1.3 dataset is a benchmark with the complex human activities for high-level video understanding, including temporal action proposal/detection and dense video captioning. There are 20,000 untrimmed long videos, with each has multiple annotated events with starting and ending time as well as the associated caption. It contains 10,024 videos for training, 4,926 for validation, and 5,044 for testing. As the ground-truth annotations of the test split are for temporal action proposal task instead of video captioning, we  simply validate our model on the validation split.

\textbf{MSVD.}
There are 1,970 short video clips collected from YouTube, with each one depicts a single activity in 10 seconds to 25 seconds. Each clip has roughly 40 English descriptions. Similar to the prior work~\cite{pan2016jointly,yao2015describing}, we take 1200 video clips for training, 100 clips for validation and 670 clips for testing.

\subsection{Implementation Details}
For the sentences in the benchmark datasets motioned above, we first remove the punctuation and convert all words into lowercase. The sentences are truncated at 28 words and tokenized. The size of word embedding size for each word is set to 468.
The POS tags of words in ground-truth are processed by Stanford Log-linear Part-Of-Speech Tagger~\cite{toutanova2003feature}, which are then divided into 14 categories for training the POS sequence generator: verb(VERB), noun(NOUN), adjective(ADJ), adverb(ADV), conjunction(CONJ), pronoun(PRON), preposition(PREP), article(ART), auxiliary(AUX), participle(PRT), number qualifier(NUM), symbol(SYM), unknown(UNK) and the end-of-sentence (EOS). Each of them also corresponds to an embedding vector with 468 dimensions.

For videos, we use TVL1-flow~\cite{perez2013tv} to compute the optical flows in both horizontal and vertical directions for adjacent frames. Then an Inflated 3D ConvNet (I3D)~\cite{carreira2017quo} trained on Kinetics action classification dataset~\cite{kay2017kinetics} extracts 8 1024-dimensional feature vectors representing the motion  features for each continuous 64 optical flow frames.
To extract the content features, we feed static frames to Inception-Resnet-v2~\cite{szegedy2017inception}, which is pre-trained on ILSVRC-2012-CLS image classification dataset~\cite{russakovsky2015imagenet}, and obtain a 1536-dimensional feature for each frame.
We also extract the spatiotemporal features by C3D network~\cite{tran2015learning}.
We take equally-spaced 30 features of a video, respectively, and pad them with zero vectors if the number of features is less than 30.

In our model, all LSTMs have a 512-dimensional hidden size, while the input dimension of LSTMs in video encoder is 1536 and 1024, which are equal to the size of content features and motion features, respectively. The input size of LSTMs in POS sequence generator and the first layer in description generator are set to 468 . The input size of the second layer of description generator is 512.

During the training, the model is optimized by the AdaDelta~\cite{zeiler2012adadelta}.
When no better CIDEr score appears in the following 30 successive validations, the training stops and the optimal model is obtained.
In the testing, we use the beam search with size 5 for the final description generation.

\subsection{Performance Comparisons}
\begin{table*}[htbp!]
\vspace{-15pt}
% \scriptsize
%   \centering
  \begin{center}
  \begin{adjustbox}{width=0.75\width}
  \begin{tabular}{c|cccccccc}
  \hline
 Training Strategy  &Model        &B@1  & B@2  &B@3  &B@4  &  M  &  R  &  C  \\ \hline \hline
 \multirow{11}{*}{Cross-Entropy}&  SA(V+C3D)~\cite{xu2016msr} &82.3 &65.7 &49.7 &36.6 &25.9 &-&- \\
            &M3(V+C3D)~\cite{wang2018m3} &73.6 &59.3 &48.3 &38.1 &26.6 &-&- \\
    % Attentional Fusion(VGG16+C3D+AUDIO)~\cite{hori2017attention} &-&-&- &39.7 &25.5 &- &40.0\\
    % Aalto(GoogleNet+C3D)~\cite{shetty2016frame}  &-&-&- &39.8 &26.9 &59.8 &45.7 \\
            &MA-LSTM(G+C3D+A)~\cite{xu2017learning} &-&-&- &36.5 &26.5 &59.8 &41.0 \\
            %&hLSTMat(R-50)~\cite{song2017hierarchical} &-&-&- &38.3 &26.3 &- &- \\
            &VideoLab(R-152+C3D+A+Ca)~\cite{ramanishka2016multimodal} &-&-&-&39.1&27.7&60.6&44.1\\
            &v2t\_navigator(C3D+A+Ca)~\cite{jin2016describing} &-&-&- &42.6 &28.8 &61.7 &46.7 \\
            &M\&M-TGM(IR+C3D+A)~\cite{chen2017video} &-&-&- &44.3 &29.4 &- &49.3 \\ 
            &{SibNet-DL(G)}~\cite{liu2018sibnet} &-&-&- &39.4 &26.9 &59.6 &45.3 \\
            &{SibNet(G)}~\cite{liu2018sibnet} &-&-&- &40.9 &27.5 &60.2 &47.5 \\
            &{MGSA(IR+C3D)}~\cite{chen2019motion} &-&-&- &42.4 &27.6 &- &47.5 \\
            &{MGSA(IR+C3D+A)}~\cite{chen2019motion} &-&-&- &45.4 &28.6 &- &50.1 \\ \hline
  \multirow{1}{*}{Reinforcement Learning} &PickNet(R-152+Ca)~\cite{chen2018less} &-&-&- &41.3 &27.7 &59.8 &44.1 \\ \hline
            %&HRL(R-152)~\cite{wang2018video} &-&-&- &41.3 &28.7 &61.7 &48.0 \\  \hline
  \multirow{5}{*}{Cross-Entropy}  &Ours(C3D+M)  &78.5 &65.3 &52.6 &41.2 &27.7 &60.9  &46.7  \\
            &Ours(I3D+M)    &79.3 &65.8 &53.3 &41.7     &27.8     &61.2  &48.5   \\
            &Ours(IR+C3D)    &79.2 &66.5 &53.7  &42.3     &28.1  &61.3 &48.6  \\
            &Ours(IR+I3D)    &79.1 &66.0 &53.3  &42.0     &28.1  &61.1 &49.0    \\
            &Ours(IR+M)    &78.4 &66.1 &53.4 &42.0     &28.2  &61.6 &48.7    \\ \hline
    \multirow{1}{*}{Reinforcement Learning} &Ours\_RL(IR+M) &81.2 &67.9 &53.8 &41.3 &28.7 &62.1 &53.4 \\ \hline
  \end{tabular}
  \end{adjustbox}
  \end{center}
  \vspace{-7pt}
  \caption{Performance comparisons with different competing models on the testing set of the MSR-VTT in terms of BLEU@1$\sim$4, METEOR, and ROUGE-L, CIDEr scores (\%). V, G, C3D, R-\textit{N}, IR, I3D, A and Ca denote VGG19, GoogleNet, C3D, \textit{N}-layer ResNet, Inception\_ResNet-v2, I3D, Audio, and Category features, respectively. M denotes the motion features from optical flow extracted by I3D.}
  \label{table:msrvtt_compare_with_ALL_competitors}
  \vspace{-15pt}
\end{table*}

In this subsection, we compare our method with the state-of-the-art methods with multiple features on benchmark datasets, including SA~\cite{yao2015describing}, M3~\cite{wang2018m3}, v2t\_navigator~\cite{jin2016describing}, Aalto~\cite{shetty2016frame}, VideoLab~\cite{ramanishka2016multimodal}, MA-LSTM~\cite{xu2017learning}, M\&M-TGM~\cite{chen2017video}, PickNet~\cite{chen2018less}, LSTM-TSA$_{IV}$~\cite{pan2017video}, {SibNet~\cite{liu2018sibnet}, MGSA~\cite{chen2019motion}}, and SCN-LSTM~\cite{Gan_2017_CVPR}, most of which fuse different features by simply concatenating.
%\textcolor{red}{Although MA-LSTM proposed a multi-level attention mechanism for multi-modal and MGSA employed the motion modality to dynamically summarize the spatial modality, which differ from the other methods, they did not consider the interaction in these modalities.}

We first show the quantitative results on MSR-VTT in Table~\ref{table:msrvtt_compare_with_ALL_competitors}. Since the methods use different CNN features or combinations of features, it is hard to make an absolutely fair comparison. Trained by cross-entropy loss, our proposed method is better than most of the competing models, including PickNet trained by RL, which demonstrates the benefits of the proposed gated fusion network as well as the incorporated global syntactic structure POS information. It is worth noticing that our model performs inferiorly to M\&M-TGM and v2t\_navigator on some metrics, such as CIDEr and METEOR. The reason is that the features of some modalities are of great differences with different expressive abilities. For example, the v2t\_navigator applies audio (A) and topic (Ca) features providing by MSR-VTT, while M\&M-TGM use a multi-task to predict the topic (Ca) features. These modalities can provide strong prior knowledge for captioning generation, which can however not be directly obtained by content or motion features. Better performances are expected if these strong features are also fused in our methods.
{Besides, same modalities are utilized in MGSA(IR+C3D) and Ours(IR+C3D), based on which our method performs better, mainly attributed to the proposed cross gating strategy and the introduced POS information.}

We also train our model by RL, specifically the self-critical sequence training~\cite{rennie2017self}, which is denoted as Ours\_RL(IR+M) in Table~\ref{table:msrvtt_compare_with_ALL_competitors}. Obviously, self-critical strategy generates  better performances than training with the traditional cross-entropy loss on all metrics except BLEU@4. It is reasonable as we mainly focus on optimizing the CIDEr metric with RL. Comparing with other competing models, Ours\_RL(IR+M) obtains the state-of-the-art performances on both ROUGE-L and CIDEr, which achieve 62.1 and 53.4, respectively. The superior performances further demonstrate the benefits of the proposed gated fusion network and the incorporation of global POS information.

Besides, we also verify our work on MSVD as shown in Table~\ref{table:msvd_compare_with_ALL_competitors}. Once again, our methods outperforms other competitors, especially on CIDEr. Without self-critical, Ours(IR+M) has obtained superior scores. 
When trained with self-critical, Ours\_RL(IR+M) significantly improves all the metric scores and achieves the new state-of-the-art results on BLEU@4, METEOR, ROUGE-L and CIDEr. 
%It demonstrates that our method is effective on different datasets.
{It is worth noticing that SibNet(G) is an excellent method that achieves the state-of-the-art on MSVD using only GoogleNet feature, as additional content and semantic branches are introduced in SibNet and significantly boost the performances. Trained with the decoder loss alone, the SibNet-DL(G) performs slightly inferiorly to our method. Our model is orthogonal to SibNet, which can be incorporated into SibNet for further boosting the performances. }

\begin{table}[htbp]
% \scriptsize
%   \centering
  \begin{center}
  \begin{adjustbox}{width=0.65\width}
  \begin{tabular}{ccccc}
  \hline
  Model        & B@4  &  M  &  R  &  C  \\ \hline \hline
    MA-LSTM(G+C3D)  &52.3  &33.6  &-  &70.4  \\
    LSTM-TSA$_{IV}$(V+C3D) &52.8  &33.5  &-  &74.0  \\
    SCN-LSTM(R-152+C3D) &50.2 &33.4 &- &77.0  \\
    M\&M-TGM(IR+C3D+A)  &48.8 &34.4 &- &80.5  \\ 
    {SibNet-DL(G)} &51.9 &33.1 &69.9 &81.9 \\
    {SibNet(G)} &\textbf{54.2} &34.8 &71.7 &88.2 \\ \hline
    Ours(IR+M)  &52.5  &34.1  &71.3  &88.7  \\
    Ours\_RL(IR+M)  &53.9  &\textbf{34.9}  &\textbf{72.1}  &\textbf{91.0}  \\ \hline
  \end{tabular}
  \end{adjustbox}
  \end{center}
  \vspace{-7pt}
  \caption{Performance comparisons with different baseline methods on the testing set of the MSVD dataset (\%).}% in terms of BLEU@4, METEOR, ROUGE-L, and CIDEr scores (\%).}
  \label{table:msvd_compare_with_ALL_competitors}
  \vspace{-7pt}
\end{table}

\begin{table}[htbp]
% \scriptsize
%   \centering
  \begin{center}
  \begin{adjustbox}{width=0.7\width}
  \begin{tabular}{ccccc}
  \hline
  Model        & B@4  &  M  &  R  &  C  \\ \hline \hline
    % EncDec(IS)    &  39.6   &   26.8   &   59.3    & 44.8   \\
    % EncDec(I3S)    &  40.3   &   27.1   &   59.5    & 45.8   \\
    % EncDec(I3F)    &  37.7   &   25.9   &   58.6    & 40.8   \\ \hline \hline
    % EncDec+F(IR+M)    &  40.4   &   27.1   &   59.6    & 47.4   \\
    EncDec+F(IR+M)    &  39.7  &   26.8   &   59.3    & 45.4   \\
    % EncDec+L(IS+I3F)    &  41.2   &   27.8   &   60.5    & 47.7   \\
    EncDec+CG(IR+M)    &  41.7   &    27.9  &   61.0    & 48.4   \\
    Ours(IR+M)    &  \textbf{42.0}   &   \textbf{28.2}   &   \textbf{61.6}    & \textbf{48.7}   \\ 
    \hline
    % EncDec+F(I3D+M)    &  40.4   &   26.9   &   59.8    & 46.9   \\
    EncDec+F(I3D+M)    &  39.7   &   26.6   &   58.8    & 45.1   \\
    % EncDec+L(I3S+I3F)    &  41.1   &   27.4   &   60.9    & 47.6   \\
    EncDec+CG(I3D+M)    &  41.4   &    27.7  &   61.0    & 47.7   \\
    Ours(I3D+M)    &  \textbf{41.7}   &   \textbf{27.8}   &   \textbf{61.2}    & \textbf{48.5}   \\   \hline
  \end{tabular}
  \end{adjustbox}
  \end{center}
  \vspace{-7pt}
  \caption{Performance comparisons with different baseline methods on the testing set of the MSR-VTT dataset(\%).  Methods of the same name but different text in the brackets indicates the same method with different feature inputs.} % in terms of BLEU@4, METEOR, and ROUGE-L, CIDEr scores 
  \label{table:msrvtt_compare_with_ours}
  \vspace{-14pt}
\end{table}

\subsection{Ablation Studies}
To demonstrate the effectiveness of the proposed components, we design several baseline models with different structures by removing certain components, which are listed as follows:
\vspace{-7pt}
\begin{itemize}
% \vspace{-5pt}
    %\item EncDec: This is the basic model where the video encoder only take motion or content features as input and does not consider any fusion strategies. Also, no POS information is utilized here.
    %\item EncDec+E: With the similar architecture as the EncDec, this model takes both motion and content features as input and concatenates them as one feature by a simple early fusion strategy.
    %\item EncDec+L: The video encoder is realized in a two-stream architecture to extract the motion and content features independently. After that, a late fusion but without cross gating is employed for merging them together.
    %\item EncDec+CG: This model makes the late fusion of EncDec+L works with the proposed cross gating block to effectively fuse the motion and content features.
    \item EncDec+F: This is the basic model where the video encoder fuses diverse features as one video representation by simple concatenating.
    \vspace{-7pt}
    \item EncDec+CG: This model employs the proposed gated fusion network to effectively fuse different features, but without the POS sequence generator.
    \vspace{-7pt}
    \item Ours (EncDec+CG+POS): It is the proposed model, where the gated fusion network is employed and the global POS tag information generated by the proposed POS sequence generator are considered for video captioning.
\end{itemize}
\vspace{-7pt}

The ablation experimental results of aforementioned models with different components on testing split of MSR-VTT are shown in Table~\ref{table:msrvtt_compare_with_ours}. %\sout{, where IR and I3D denote content features extracted by Inception\_ResNet\_V2 and I3D, respectively, and M denotes the motion features extracted by I3D. These features are divided into two combinations, including (IR, M) pair and (I3D, M) pair, each of which consists of content  and motion features.} 
We use different feature combinations, \eg, (IR, M) pair and (I3D, M) pair, each of which consists of one content and one motion features.

Compared with the basic model EncDec+F(IR+M) that simply concatenates the features, we can observe a significant improvement on all the evaluation metrics by incorporating the proposed gated fusion network in EncDec+CG(IR+M). The similar performance improvements also appear in EncDec+CG(I3d+M). The better scores on BLUE@4, METEOR, ROUGE-L and CIDEr of EncDec+CG in Table~\ref{table:msrvtt_compare_with_ours} demonstrate that:
1) There exist complicated semantic relationships between different features, specifically the content feature and motion features, which a simple concatenation can not capture.
2) By performing the gated fusion network on the content and motion features, a more representative video feature for video captioning can be obtained.

{To further demonstrate the effectiveness of the proposed fusion strategy, we also compare the gated fusion network with more complex fusion algorithms, such as MCB~\cite{fukui2016multimodal}, MLB~\cite{kim2016hadamard}, and element-wise adding, which used compact bilinear pooling, low-rank bilinear pooling, and feature vector adding to exploit the relationship of each element in different modal features, respectively. The performance comparisons are illustrated in Table~\ref{table:msrvtt_compare_with_multi_fusion_algorithms}. The cross gating mechanism enhances the relevant part of different modalities and uses a residual structure to retain information that may not be relevant but unique, which can not be modeled by MCB, MLB, or element-wise adding. As such, the relationships between different modalities can be more comprehensively exploited to further benefit the video captioning.}

\begin{table}[tbp]
  \vspace{-0pt}
  \begin{center}
  \begin{adjustbox}{width=0.65\width}
  \begin{tabular}{ccccc}
  \hline
    Model      & B@4  &  M  &  R  &  C  \\ \hline \hline
    MCB(IR+M)  &41.2  &27.5  &60.6  &46.3  \\
    MLB(IR+M) &41.4  &27.6  &60.9  &47.6  \\
    Element-wise adding(IR+M) &40.2 &27.0 &60.2 &46.3  \\ \hline
    EncDec+CG(IR+M)  &\textbf{41.7}  &\textbf{27.9}  &\textbf{61.0}  &\textbf{48.4}  \\ \hline
  \end{tabular}
  \end{adjustbox}
  \end{center}
  \vspace{-7pt}
  \caption{{Performance comparisons with different fusion strategies on the testing set of the MSR-VTT(\%).}}
  \label{table:msrvtt_compare_with_multi_fusion_algorithms}
  \vspace{-14.5pt}
\end{table}

Our proposed method, considering both the gated fusion network and the global POS information, takes a further step on the EncDec+CG and is trained by adaptively and dynamically incorporating POS to guide the generation of each word. As illustrated in Table~\ref{table:msrvtt_compare_with_ours}, the proposed model yields the highest performance  on the four metrics with the same features pair, namely IR and M. The same observations can be observed if using (I3D, M) pair as the feature representations in Table~\ref{table:msrvtt_compare_with_ours}, which proves that the POS information provides a global view on the potential syntactic structure of its language descriptions and thereby further improves the performance of video captioning.

\subsection{Qualitative Analysis}
\begin{figure}
\centering
\includegraphics[scale=0.8]{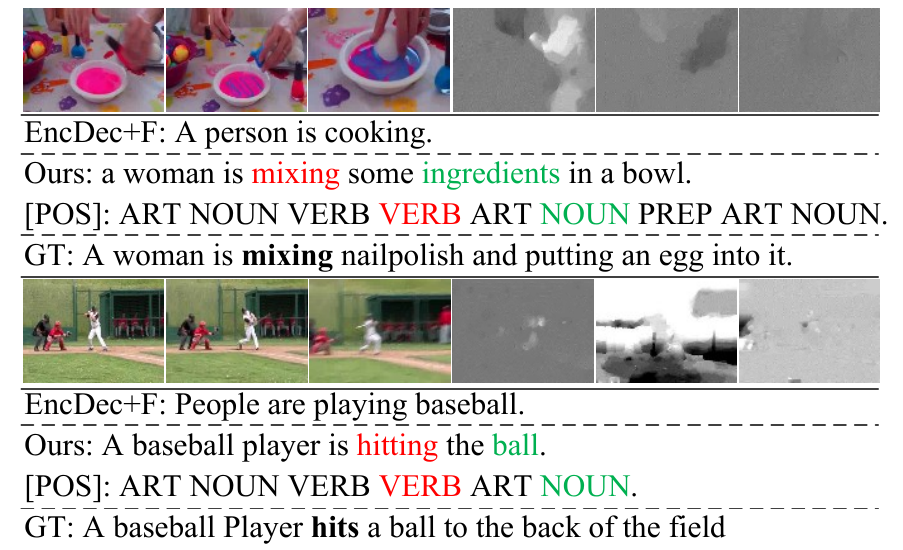} %[width=0.8\hsize]
\vspace{-3pt}
\caption{Visualization of some video captioning examples on the MSR-VTT with the basic model and the proposed model. Due to the page limit, only one ground-truth (GT) sentence is given as reference. Also we illustrate the generated  POS sequence. Compared to the base model EncDec+F, the proposed model yields more accurate sentence descriptions. %Words that generated under POS tag guidance are highlighted in red and green.
}
\label{fig:results}
\vspace{-10pt}
\end{figure}
Besides, some qualitative examples are shown in Fig.~\ref{fig:results}. it can be observed that the proposed model, with the cross gating strategy and incorporating global POS information, can generate more accurate descriptions than the baseline model. For example, in the first example our model realizes that it is not related to cooking and correctly predicts the action `\textit{mixing}' and object `\textit{ingredients}' under the guidance of POS tag `\textit{VERB}' highlighted in red and `\textit{NOUN}' highlighted in green, respectively. In the second example, compared to the general description of the EncDec+F, our model accurately captures the detail `\textit{hitting the ball}', which makes a more specific and vivid description.

\begin{figure}[htbp]
\centering
\includegraphics[scale=0.8]{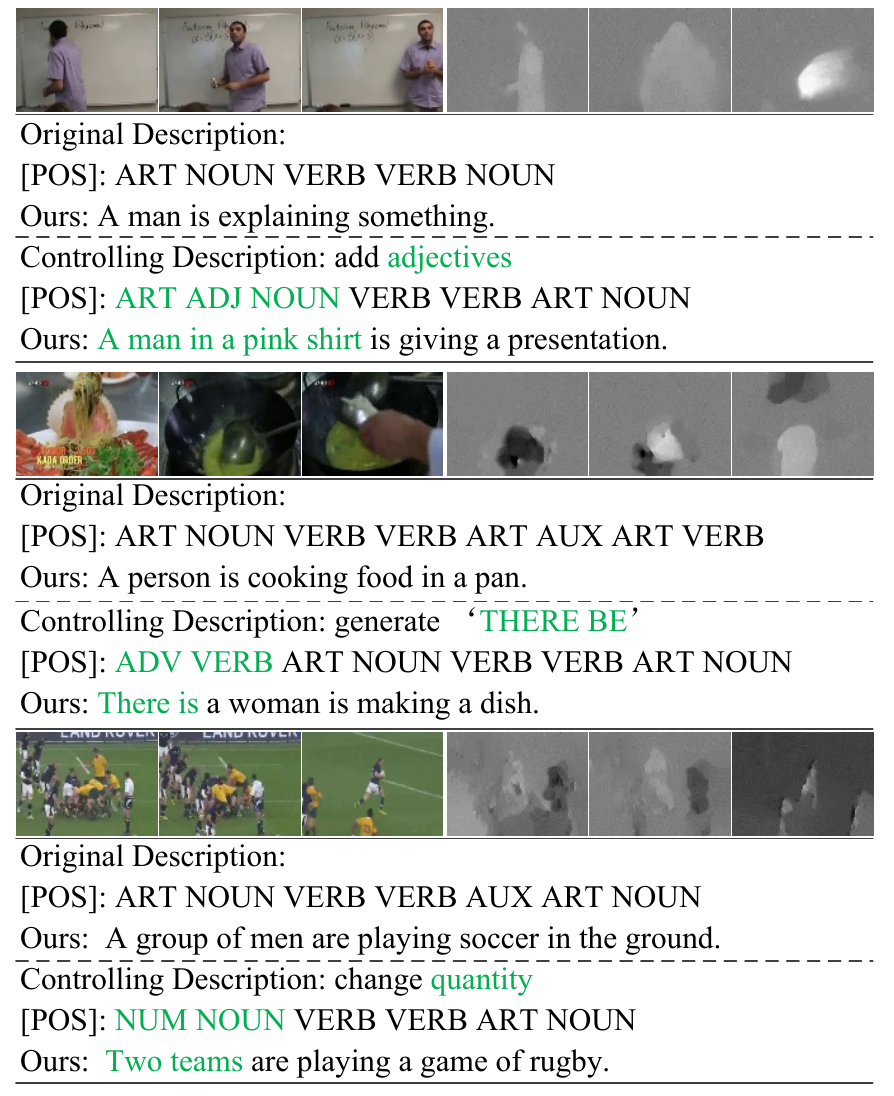} %width=\hsize
\vspace{-3pt}
\caption{Visualization of some video captioning examples on the MSR-VTT by controlling the captioning generation with modifying the generated POS tag sequence. The POS tags in green denote the human modified ones, while the words in green are generated under the guidance of the modified global POS information.
}
\label{fig:controllable}
\vspace{-10pt}
\end{figure}

\subsection{Controllable of Syntax for Video Captioning}
{Finally, we show the controllable of the syntactic structure for generating captions by manually modifying the generated POS tag sequence in the inference stage. For example, when we expect an adjective on the current time step, the predicted POS tag will be replaced by `\textit{ADJ}' tag manually, whatever the predicted result is. The changed `\textit{ADJ}' tag is subsequently fed to the POS sequence generator and the next POS tag is predicted, meanwhile the hidden state of POS sequence generator is modified. As such, the global POS information can be modified to control the overall syntactic structure of the generated sentence.}
%in a particular time step,

Some examples are illustrated in Fig.~\ref{fig:controllable}. For the first sample, we add the '\textit{ADJ}' in the front of the subject, so that we can describe the event with more details. With the changed POS information, the description generator predicts ``\textit{a man in a pink shirt}'', which is in line with our expectations. In the second sample, we would like to generate a sentence with ``\textit{there be}'' as the beginning. Our approach once again meet the requirement. The most interesting thing is when we replace the article (`\textit{ART}') with number (`\textit{NUM}'), the generator provides ``\textit{two teams}'', instead of ``\textit{two men}'', to replace ``\textit{a group of men}''. These results demonstrate that the proposed gated fusion network effectively and fully captures semantic meaning of the video by understanding the relationship between different features. Therefore, even though the global POS information is changed, it can accurately generate the reliable descriptions. Meanwhile, the global POS information can indeed control the overall syntactic structure of the generated sentence. { More experimental results can be referred to the supplementary material}\footnote{{\color{blue} \url{https://github.com/vsislab/Controllable_XGating/blob/master/supplementary.pdf}}}.
% \url{https://github.com/vsislab/Controllable_XGating/blob/master/supplementary.pdf}

\section{Conclusions}
% In this paper, we proposed a novel model for video captioning, with a gated fusion network and a POS sequence generator, which can fuses diverse information with a cross gating strategy and produce a global syntactic structure as the guidance for video captioning, respectively. The proposed model archives competitive performances on both MSR-VTT and MSVD datasets, which indicates the superiority of our proposed model. Moreover, the generated global  POS  information   can  be  further used to control the syntactic structure of the generated caption, which thereby improves the corresponding diversity.
In this paper, we proposed a novel model for controllable video captioning using a gated fusion network and a POS sequence generator. This model can fuse diverse information with a cross-gating strategy and produce a global syntactic structure as the guidance for addressing video captioning. The proposed model achieves competitive performances on both MSR-VTT and MSVD datasets, which indicates the superiority of our model. Moreover, the generated global POS information can be further leveraged to control the syntactic structure of the generated caption, thereby improving the corresponding diversity.

%oMnsoreover, our proposedapti cross gating strategy and POS sequence generator can be incorporated with any video captioning models and believed to further boost their performances.

\section*{Acknowledgments}
The authors would like to thank the anonymous reviewers for the constructive comments to improve the paper. This work was supported in part by the National Key Research and Development Plan of China under Grant 2017YFB1300205, in part by NSFC under Grant 61573222, and in part by the Major Research Program of Shandong Province under Grant 2018CXGC1503.

{\small
\bibliographystyle{ieee_fullname}
\bibliography{egbib}
}

\end{document}

%%%%%%%%% TITLE
% \title{Gated Fusion Network for Video Captioning \\ with POS Sequence Guidance  \\[10pt] ---Supplementary Material \\[10pt] Paper ID 1039} 
\title{Controllable Video Captioning  with POS Sequence Guidance \\Based on Gated Fusion Network  \\[10pt] ---Supplementary Material}

\author{Bairui Wang$^1$\thanks{This work was done while Bairui Wang was a Research Intern with Tencent AI Lab.} \qquad Lin Ma$^2$\thanks{Corresponding authors.} \qquad Wei Zhang$^{1\dagger}$ \qquad Wenhao Jiang$^2$  \qquad Jingwen Wang$^2$ \qquad Wei Liu$^2$  \\
$^1$School of Control Science and Engineering, Shandong University \qquad $^2$Tencent AI Lab \\
{\tt\small \{bairuiwong, forest.linma, cswhjiang, jaywongjaywong\}@gmail.com} \\ {\tt\small  davidzhang@sdu.edu.cn \qquad wl2223@columbia.edu}
}

\maketitle
%\thispagestyle{empty}

%%%%%%%%%%% body %%%%%%%%%%%%%%%%%
In this appendix, we add some technical details mentioned in the submitted ICCV2019 manuscript, entitled as ``Controllable Video Captioning  with POS Sequence Guidance Based on Gated Fusion Network'' with paper ID as 1039, and present extra ablation experiments on the ActivityNet 1.3~\cite{caba2015activitynet}.
Specifically, we first introduce  the self-critical sequence training~\cite{rennie2017self} method, which is employed for training our model. Then we analysis the experimental results on the ActivityNet 1.3. And Finally, more supplementary qualitative results of our model on MSR-VTT dataset are illustrated.

\section{Reinforcement Learning}
In the submitted manuscript, we intend to directly train the captioning models guided by the evaluation metrics, specifically the CIDEr~\cite{vedantam2015cider} in this work, instead of the cross-entropy losses. However, such a evaluation metric is discrete and non-differentiable, which makes the network difficult to be optimized with traditional methods. As such, we employ the reinforcement leaning (RL)~\cite{sutton1998introduction} which has been widely used in both image captioning~\cite{rennie2017self,ren2017deep} and video captioning~\cite{pasunuru2017reinforced,wang2018video}. We resort to the self-critical sequence training~\cite{rennie2017self}, an excellent REINFORCE-based algorithm, that is specializing in processing the discrete and non-differentiable variables and first purposed for boosting image captioning. 

\subsection{REINFORCE Algorithms}
In our case, the videos and words can be considered as the \textit{environment}, and the proposed captioning model is considered as the \textit{agent} that interacts with the \textit{environment}. The parameters $\theta_{gen}$ of the model define the policy $\pi_{\theta_{gen}}$ which takes an \textit{action} to predict a word followed by the updating of the \textit{state}, that is the hidden states, the memory cells, and other learnable parameters of the captioning model.

The \textit{reward} is obtained when a sentence is generated, which denotes the score of language metric CIDEr in this work. The model is trained by minimizing the negative expected reward:
\begin{align}
\label{eq:rl1}
  \begin{split}
    \mathcal{L}_{RL}(\theta_{gen}) = -\mathbb{E}_{\textbf{S}^{k} \sim \pi_{\theta_{gen}}}\left [ r \left ( \textbf{S}^k \right )  \right ],
  \end{split}
\end{align}
where $\textbf{S}^{k}$ denotes the sentence sampled by the model for the $k_{th}$ video in the dataset. Subsequently, the $r \left ( \textbf{S}^k \right )$ denotes the reward of the sentence, that is CIDEr in our work. Using the REINFORCE algorithm, the gradient of the non-differentiable reward function in Eq.~(\ref{eq:rl1}) can be obtained as follows:
\begin{align}
\label{eq:rl2}
  \begin{split}
    \triangledown _{\theta_{gen}} \mathcal{L}_{RL} \left( \theta_{gen} \right) = -\mathbb{E}_{\textbf{S}^k \sim \pi_{\theta_{gen}}} \left[ r \left( \textbf{S}^k  \right ) \cdot \triangledown_{\theta_{gen}} \log\pi_{\theta_{gen}} \left(\textbf{S}^k \right )  \right ].
  \end{split}
\end{align}

For each sample from the training set, as $\mathcal{L}_{RL} \left( \theta_{gen} \right)$ is generally estimated with a single sample from $\pi_{\theta}$, the Eq.~(\ref{eq:rl2}) can be represented as follows:
\begin{align}
\label{eq:rl3}
  \begin{split}
    \triangledown _{\theta_{gen}} \mathcal{L}_{RL} \left( \theta_{gen} \right) \approx  - r \left( \textbf{S}^k  \right ) \cdot \triangledown_{\theta_{gen}} \log\pi_{\theta_{gen}} \left(\textbf{S}^k \right ).
  \end{split}
\end{align}

However, estimating the gradient with a single sample will inevitably result in a high variance. To solve this issue, a baseline reward  $b$ is usually used to generalize the policy gradient without influencing the expected gradient, if $b$ is not a function of $\textbf{S}^k$~\cite{sutton1998introduction}, which can be represented as follows:
\begin{align}
\label{eq:rl4}
  \begin{split}
    \triangledown _{\theta_{gen}} \mathcal{L}_{RL} \left( \theta_{gen} \right) \approx  - \left(  r \left( \textbf{S}^k  \right ) - b \right) \cdot \triangledown_{\theta_{gen}} \log\pi_{\theta_{gen}} \left(\textbf{S}^k \right ).
  \end{split}
\end{align}

With the chain rule, the gradient of the loss function can also be written as:
\begin{align}
\label{eq:rl5}
  \begin{split}
    \triangledown _{\theta_{gen}} \mathcal{L}_{RL} \left( \theta_{gen} \right) =\sum_{t=1}^n \frac{\partial \mathcal{L}_{RL}\left( \theta_{gen} \right )}{\partial u_t} \frac{\partial u_t}{\partial \theta_{gen}},
  \end{split}
\end{align}
where $u_t$ denotes the item that be input to the Softmax function at the $t_{th}$ time step of the description generator, that is $\left( \mathbf{W}^{(D)}_s h^{(D2)}_{t} + \mathbf{b}^{(D)}_s \right)$ in Eq.~(9) of the submitted manuscript. Using REINFORCE, the estimate of the gradient of $\frac{\partial \mathcal{L}_{RL}\left( \theta_{gen} \right )}{\partial u_t}$ with the baseline is given by~\cite{zaremba2015reinforcement} as follows:
\begin{align}
\label{eq:rl6}
  \begin{split}
    \frac{\partial \mathcal{L}_{RL}\left( \theta_{gen} \right )}{\partial u_t} \approx \left( r \left( \mathbf{S}^k \right) - b \right) \left( \pi_{\theta_{gen}} \left( s^k_t \right) - 1_{s^k_t} \right),
  \end{split}
\end{align}
where $s^k_t$ denotes the $t_{th}$ word in the sentence for the $k_{th}$ video.

\subsection{Self-critical Sequence Training}
Based on REINFORCE algorithm, the self-critical sequence training is first proposed by Rennie~\textit{et al.} for image captioning, and greatly improves the performance~\cite{rennie2017self}. 

Instead of learning another reward network for estimating the baseline reward $b$, Rennie~\textit{et al.} utilize the reward obtained by the current model under the inference algorithm in testing stage as the baseline reward $b$, that is $b=r\left( \hat{\mathbf{S}} \right)$, where $\hat{\mathbf{S}}$ is the sentence generated with the greedy search strategy by the current model. Thus, the estimate of the gradient in Eq.~(\ref{eq:rl6}) can be rewritten as:
\begin{align}
\label{eq:rl7}
  \begin{split}
    \frac{\partial \mathcal{L}_{RL}\left( \theta_{gen} \right )}{\partial u_t} \approx \left( r \left( \mathbf{S}^k \right) - r\left( \hat{\mathbf{S}} \right) \right) \left( \pi_{\theta_{gen}} \left( s^k_t \right) - 1_{s^k_t} \right).
  \end{split}
\end{align}

From aforementioned description, it can be observed that if a sample policy results a higher $r\left( \mathbf{S}^k \right)$ than the baseline $r\left( \hat{\mathbf{S}} \right)$, such a policy is encouraged by increasing the probability of the corresponding word. Conversely, those sample strategies with low rewards will be suppressed. The self-critical sequence training reduces the gradient variance as well as trains the model more effectively as only a forward propagation is required for the baseline estimating.

\section{Experiments on ActivityNet 1.3}
\subsection{ActivityNet 1.3}
The ActivityNet 1.3 dataset~\cite{caba2015activitynet} is a large scale benchmark with the complex human activities for high-level video understanding, including temporal action proposal, action detection, and dense video captioning. There are 20,000 untrimmed long videos, with each has multiple annotated events with starting and ending time as well as the associated caption. It contains 10,024 videos for training, 4,926 for validation, and 5,044 for testing. 
For the training and validation data, we construct the video-sentence pairs by extracting the labeled video segments indicated by the starting and ending time stamps, as well as their associated sentences. 
As the ground-truth annotations of the testing split are for temporal action proposal task instead of video captioning, we simply validate our model on the validation split.

\subsection{Ablation Studies}
%There are few methods utilize diverse features to validate the performance on the validation split of ActivityNet 1.3, thereby it is meaningless to compare our method with existing methods (Obviously, multiple features have advantages over a single feature). 
We conduct the ablation studies on ActivityNet 1.3 to further verify the effectiveness and reliability of our proposed model. The ablation experimental results are shown in Table.~\ref{tab:tab1}.
\begin{table}[htbp]
\begin{center}
\begin{tabular}{c|c|c|c|c}
\hline
Model & B@4  &M  &R  &C \\
\hline\hline
EncDec+F (IR+M) &4.4 &9.5  &20.2  &32.4 \\
EncDec+CG (IR+M) &4.9  &9.8  &22.4  &36.7 \\
Ours (IR+M) &\textbf{5.0}  &\textbf{10.2}  &\textbf{22.9}  &\textbf{37.3}  \\
\hline\hline
EncDec+F (I3D+M) &4.6 &9.6  &21.0  &34.3 \\
EncDec+CG (I3D+M) &\textbf{5.0}  &10.2  &22.9  &37.2 \\
Ours (I3D+M) &\textbf{5.0}  &\textbf{10.3}  &\textbf{23.0}  &\textbf{38.4}  \\
\hline
\end{tabular}
\end{center}
\caption{Performance comparisons with different baseline models on the validation split of ActivityNet 1.3 in terms of BLEU@4 (B@4), METEOR (M), ROUGE-L (R), and CIDEr (C) scores (\%). Methods of the same name but different text in the brackets indicates the same method with different feature inputs. IR and I3D denote the visual content features extracted from RGB frames by Inception\_ResNet\_V2 and I3D networks, respectively, and M denotes the motion feature extracted from optical flows by I3D network.}
\label{tab:tab1}
\end{table}

Taking IR and M as inputs, we find that the EncDec+CG (IR+M) outperforms the EncDec+F (IR+M) on all the evaluation metrics. For example, EncDec+CG (IR+M) is 4.3\% higher than EncDec+F (IR+M) on CIDEr score, which is a significant improvement on ActivityNet 1.3.
Once again, it demonstrates the proposed gated fusion network can particularly explore the relationships between different features, \eg, the IR and M in this experiment, and merge them in an effective way. While compare ours (IR+M) with EncDec+CG (IR+M), a further improvement brought by global POS information is observed. 
The similar performance improvements can also be obtained when taking I3D and M as inputs.
It indicates that the proposed gated fusion network for video captioning with the global POS guidance is effectively and generalizable.

\newpage
\section{More Qualitative Samples}

\begin{figure}[h]
\vspace{20pt}
\begin{center}
\includegraphics[scale=0.9]{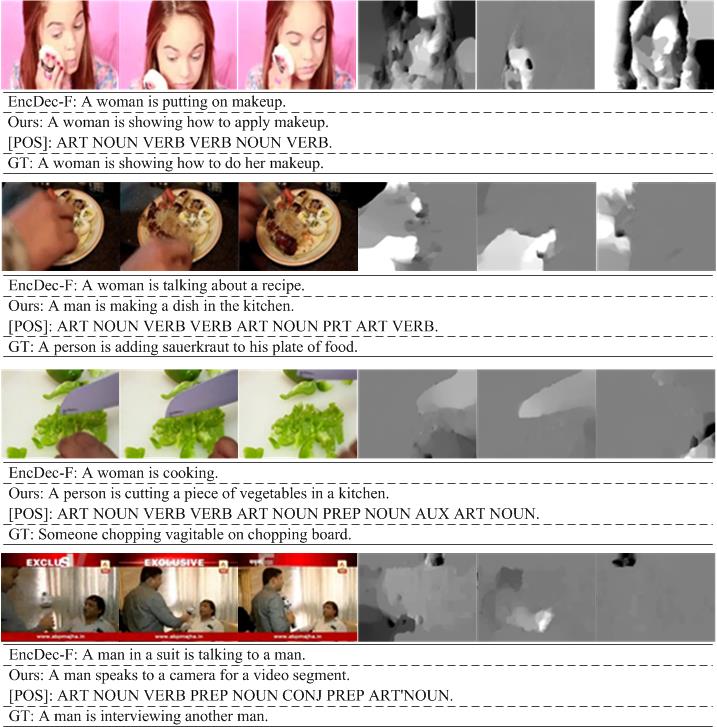}
\end{center}
\end{figure}

\newpage
\begin{figure}[h]
\vspace{60pt}
\begin{center}
\includegraphics[scale=0.9]{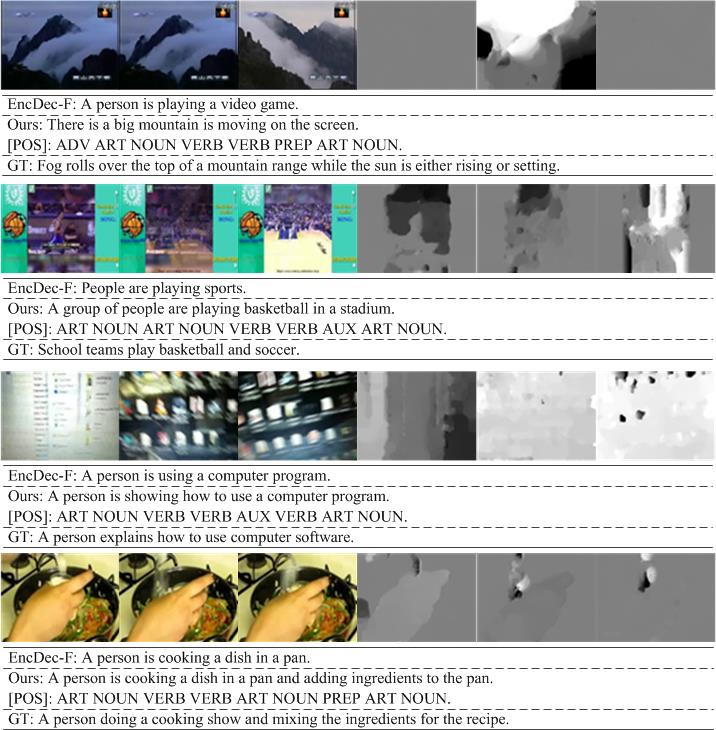}
\end{center}
\end{figure}

\newpage
\begin{figure}[h]
\vspace{60pt}
\begin{center}
\includegraphics[scale=0.9]{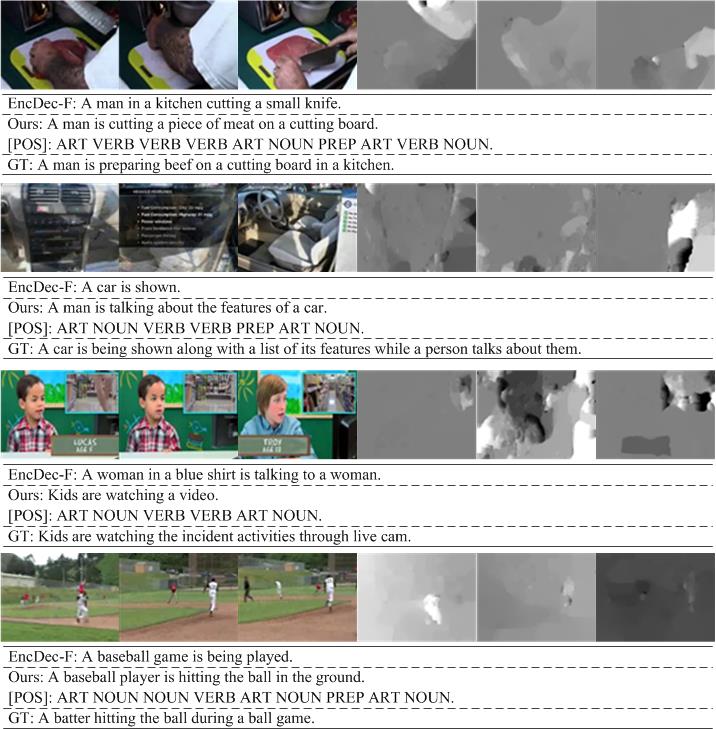}
\end{center}
\vspace{15pt}
\end{figure}

\newpage
\section{More Controllable Samples}

\begin{figure}[h]
\vspace{40pt}
\begin{center}
\includegraphics[scale=0.9]{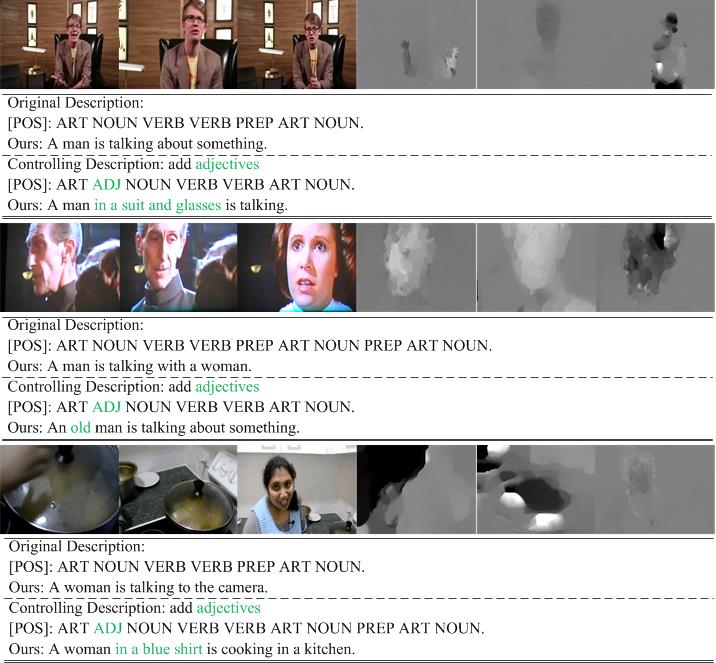}
\end{center}
\end{figure}

\newpage
\begin{figure}[h]
\begin{center}
\includegraphics[scale=0.9]{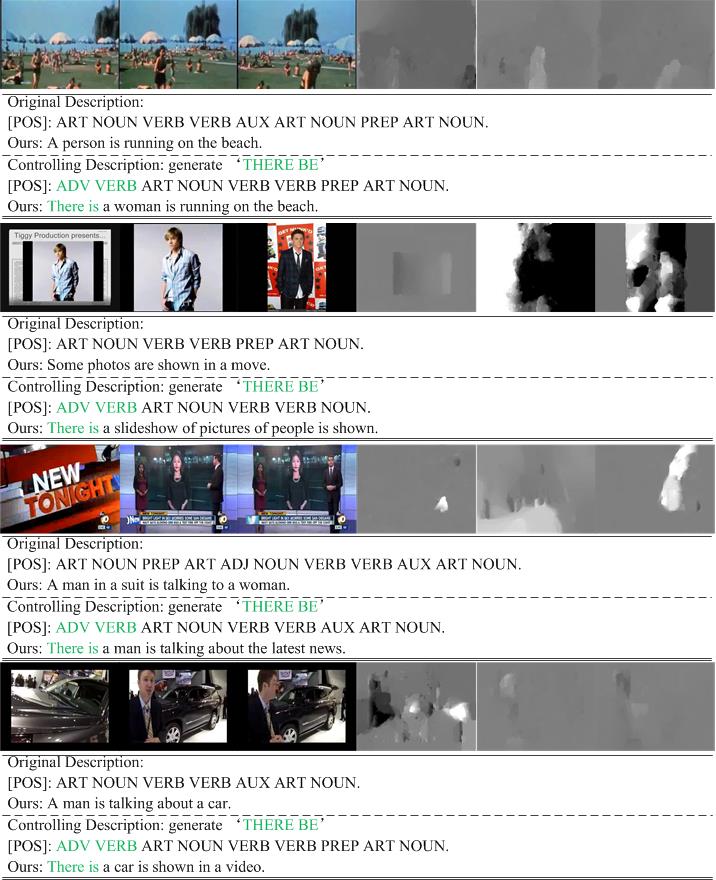}
\end{center}
\end{figure}

\newpage
\begin{figure}[h]
\begin{center}
\includegraphics[scale=0.9]{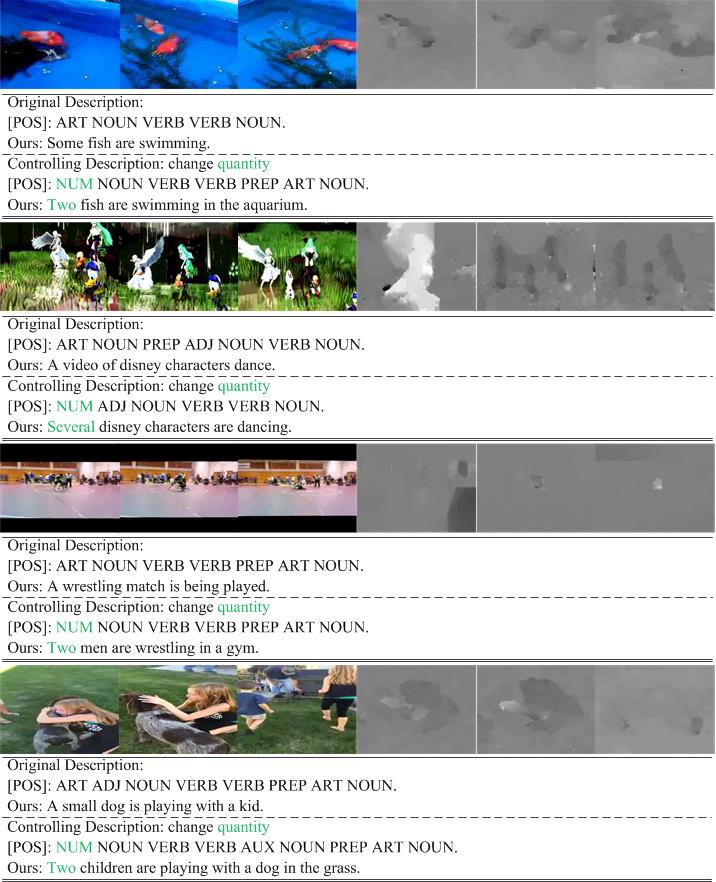}
\end{center}
\end{figure}

\newpage
{\small
\bibliographystyle{ieee}
\bibliography{egbib_supp}
}